\documentclass[10pt,twocolumn,letterpaper]{article}

\usepackage{iccv}              
\usepackage{tikz}

\definecolor{ForestGreen}{RGB}{34,139,34}
\usepackage{multirow}
\usepackage{arydshln}
\usepackage[skip=3pt]{caption}
\definecolor{iccvblue}{rgb}{0.21,0.49,0.74}
\definecolor{moderate}{rgb}{0.85, 0.55, 0.1} 
\definecolor{BrightRed}{rgb}{0.85, 0.25, 0.25}
\definecolor{ViridianGreen}{rgb}{0.25, 0.6, 0.45}

\usepackage[pagebackref,breaklinks,colorlinks,allcolors=iccvblue]{hyperref}
\usepackage{booktabs}
\usepackage{arydshln}
\usepackage{pifont}
\usepackage{xcolor}
\newcommand{\cmark}{\ding{51}}%
\newcommand{\xmark}{\ding{55}}%

\def\paperID{7620} 
\def\confName{ICCV}
\def\confYear{2025}

\title{Long-Context State-Space Video World Models}

\author{
Ryan Po\textsuperscript{1} \quad
Yotam Nitzan\textsuperscript{3} \quad
Richard Zhang\textsuperscript{3} \quad
Berlin Chen\textsuperscript{2} \quad
Tri Dao\textsuperscript{2} \\
Eli Shechtman\textsuperscript{3} \quad
Gordon Wetzstein\textsuperscript{1} \quad
Xun Huang\textsuperscript{3} \\
\textsuperscript{1}Stanford University \quad
\textsuperscript{2}Princeton University \quad
\textsuperscript{3}Adobe Research
}

\begin{document}
\maketitle
\begin{abstract}
Video diffusion models have recently shown promise for world modeling through autoregressive frame prediction conditioned on actions. However, they struggle to maintain long-term memory due to the high computational cost associated with processing extended sequences in attention layers. To overcome this limitation, we propose a novel architecture leveraging state-space models (SSMs) to extend temporal memory without compromising computational efficiency. Unlike previous approaches that retrofit SSMs for non-causal vision tasks, our method fully exploits the inherent advantages of SSMs in causal sequence modeling. Central to our design is a block-wise SSM scanning scheme, which strategically trades off spatial consistency for extended temporal memory, combined with dense local attention to ensure coherence between consecutive frames. We evaluate the long-term memory capabilities of our model through spatial retrieval and reasoning tasks over extended horizons. Experiments on Memory Maze and Minecraft datasets demonstrate that our approach surpasses baselines in preserving long-range memory, while maintaining practical inference speeds suitable for interactive applications.
\end{abstract}
\vspace{-24pt}
\section{Introduction}
\label{sec:intro}

World models~\cite{ha2018recurrent,hafner2020dream,wu2023daydreamer,generalworldmodelsurvey,micheli2023transformers,hu2023gaia,wang2023world,Genesis} are causal generative models designed to predict how world states evolve in response to actions, enabling interactive simulation of complex environments.
Video diffusion models~\cite{ho2022video,videoworldsimulators2024,polyak2024movie,yang2025cogvideox,kong2024hunyuanvideo,kang2024how,ma2025step} have emerged as a promising approach for world modeling.
While early models were limited to generating fixed-length videos and therefore unsuitable for interactive applications, recent architectures have enabled infinite-length video generation through autoregressive, sliding-window prediction~\cite{weng2024art,parkerholder2024genie2,yin2024causvid,gao2024ca2,oasis2024,hu2024acdit,jin2024pyramidal,feng2024matrix,valevski2024diffusion,alonsodiffusion,yu2025gamefactory,he2025pre,che2024gamegen,song2025history}.
This paves the way for a new paradigm in which video diffusion models can interactively simulate visual worlds by continuously generating video frames conditioned on interactive control signals.

\begin{figure}[t!]
    \centering
    \includegraphics[width=\linewidth]{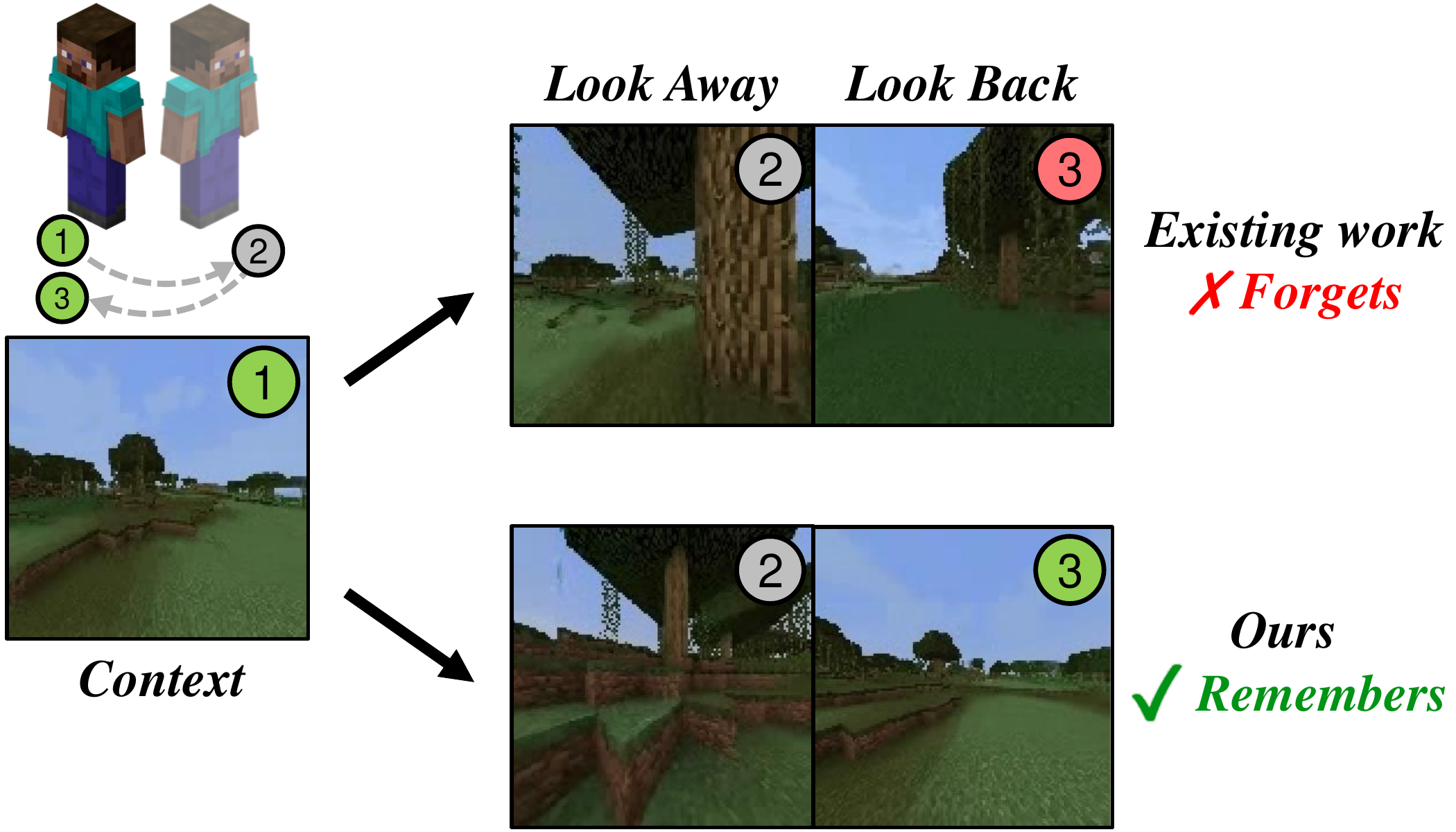}
    \caption{\textbf{Failure case of existing video world models.} Without long-term memory, previously observed regions may appear altered or inconsistent upon revisiting. 
    }
    \label{fig:motivation}
    \vspace{-6pt}
\end{figure}

\begin{table*}[t!]
    \centering
    \small
    \begin{tabular}{lcccc}
        \toprule
        
        \multirow{2}{*}{\textbf{Architecture}} & \multirow{2}{*}{\shortstack[c]{\textbf{Training} \\ \textbf{Complexity}}} & \multicolumn{2}{c}{\textbf{AR Inference}} & \multirow{2}{*}{\shortstack[c]{\textbf{Long} \\ \textbf{Memory}}} \\ \cline{3-4} 
        & & \footnotesize{\textbf{(in total)}} & \footnotesize{\textbf{(per frame)}} & \\
        \midrule
        Bidirectional attention & \textcolor{BrightRed}{Quadratic} & \textcolor{BrightRed}{Cubic} & \textcolor{BrightRed}{Quadratic} & \textcolor{ForestGreen}{\cmark} \\
        Causal attention & \textcolor{BrightRed}{Quadratic} & \textcolor{moderate}{Quadratic} & \textcolor{moderate}{Linear} & \textcolor{ForestGreen}{\cmark} \\
        Causal + sliding window inference & \textcolor{moderate}{Sub-quadratic} & \textcolor{ForestGreen}{Linear}& \textcolor{ForestGreen}{Constant} & \textcolor{BrightRed}{\xmark} \\ 
        \midrule
        Ours (SSM + local causal attention) & \textcolor{ForestGreen}{Linear} & \textcolor{ForestGreen}{Linear} & \textcolor{ForestGreen}{Constant} & \textcolor{ForestGreen}{\cmark} \\
        \bottomrule
    \end{tabular}
    \caption{Computational complexity comparison for autoregressive video diffusion architectures, with respect to sequence length. Bidirectional attention models process all previous frames with quadratic complexity when generating a single frame, while causal attention models improve efficiency through KV-caching but still scale linearly with video length.
    Both approaches commonly employ sliding-window inference to maintain manageable computational complexity.
    While sliding window inference with causal attention achieves constant per-frame inference time, it sacrifices long-term memory.
    Our hybrid architecture combines SSMs with local causal attention, maintaining long-term memory while achieving constant per-frame inference speed—ideal for interactively generating a persistent world. 
    }
    \label{tab:model-comparison}
    \vspace{-4pt}
\end{table*}

However, existing video world models have very limited temporal memory due to the restricted context length of their attention mechanisms.
This limitation hinders their ability to simulate a persistent world with long-term consistency. For example, when using an existing video world model to simulate a game, the entire environment might completely change after a player simply looks right and then left again (see Fig.~\ref{fig:motivation}). This is because the frames that contain the original environment are no longer in the model's attention window.
While one could theoretically extend the memory by training the model on longer context windows, this approach faces two major limitations: (1) the computational cost of training scales quadratically with context length, making it prohibitively expensive, and (2) the per-frame inference time grows linearly with context length, resulting in increasingly slower generation speed that would be impractical for applications that require real-time, infinite-length generation~(such as gaming).

In this work, we propose a novel video world model architecture that leverages state-space models (SSMs) to enable long-term memory while maintaining efficient training and fast autoregressive inference.
Our key innovation is a block-wise scan scheme of Mamba~\cite{gumamba} that optimally balances temporal memory and spatial coherence, while preserving temporal causality.
We complement this with dense local attention between neighboring frames, maintaining high-fidelity generation with minimal computational overhead.
Unlike previous methods that retrofit SSMs for non-causal vision tasks, our approach fundamentally differs by specifically employing SSMs to handle causal temporal dynamics and track the state of the world, fully exploiting their inherent strengths for sequence modeling.
Tab.~\ref{tab:model-comparison} compares our method with existing solutions. Unlike architectures with full bidirectional~\cite{che2024gamegen,song2025history,yu2025gamefactory} or causal attention~\cite{parkerholder2024genie2,yin2024causvid,gao2024ca2,oasis2024,hu2024acdit,jin2024pyramidal,feng2024matrix}, our model achieves constant per-frame inference time, making it particularly suitable for interactive applications that require infinite rollout.
Furthermore, our approach maintains consistent long-term memory, unlike existing methods that apply causal attention within a sliding window~\cite{yin2024causvid,oasis2024}. 

The remainder of this paper is structured as follows: Our approach builds on autoregressive video diffusion models trained with independent per-frame noise levels~\cite{chen2024diffusion}, as well as SSM architectures. Sec.~\ref{sec:background} provides the necessary technical background on these components. In Sec.~\ref{sec:method}, we detail our novel architecture design that combines block-wise SSM scanning with local causal attention, along with our specialized training and sampling strategies. Sec.~\ref{sec:experiments} introduces evaluation metrics that are used to assess long-term memory in video world models.
Specifically, the spatial retrieval metric quantifies the model's ability to retrieve the environment when revisiting a previously observed location, while the spatial reasoning metric requires the model to infer the appearance at unvisited locations based on past observations from different viewpoints.
Experiments demonstrate that our approach significantly outperforms baseline architectures under these metrics in challenging Memory Maze and Minecraft datasets, confirming its effectiveness in maintaining long-term memory.






\section{Related Work}
\label{sec:related_work}

\paragraph*{Video generation.}
Modern video generative models rely on either autoregressive~(AR) prediction of discretized tokens~\cite{sun2024autoregressive,kondratyuk2024videopoet,yan2021videogpt,wang2024loong} or diffusion models. Compared to discretized AR models, diffusion models usually produce higher-quality videos.
They first demonstrate remarkable success in image synthesis~\cite{rombach2022high,ramesh2022hierarchical} and have been extended to videos by treating the temporal dimension analogously to spatial dimensions, generating the entire video of a fixed length in a single process~\cite{blattmann2023stable,li2022efficient,ho2022video,singer2022make,yang2024cogvideox,kong2024hunyuanvideo,hacohen2024ltx, oshima2024ssm}. 
%

Although effective for short clips, diffusion models face significant limitations when scaling to longer video sequences due to prohibitive computational demands. Additionally, they lack support for online, incremental generation of videos.
%
%
To address this, researchers have explored training strategies for video diffusion models that enable AR inference.
Some works train conditional diffusion models to denoise a few next frames conditioned on past clean frames~\cite{ho2022video,he2022latent,chen2023seine,jin2024pyramidal,zhang2024extdm,gao2024ca2,gao2024vid}, while others introduce per-frame independent noise levels during training which also enable AR inference~\cite{chen2024diffusion,yin2024causvid,song2025history}.
%
%
At inference time, both approaches can generate long videos autoregressively, typically through a sliding-window mechanism due to computational constraints.
%
This inherently restricts long-term memory and prevents the model from maintaining temporal consistency over extended sequences.

\begin{figure*}[t]   
    \centering
    \includegraphics[width=\linewidth]{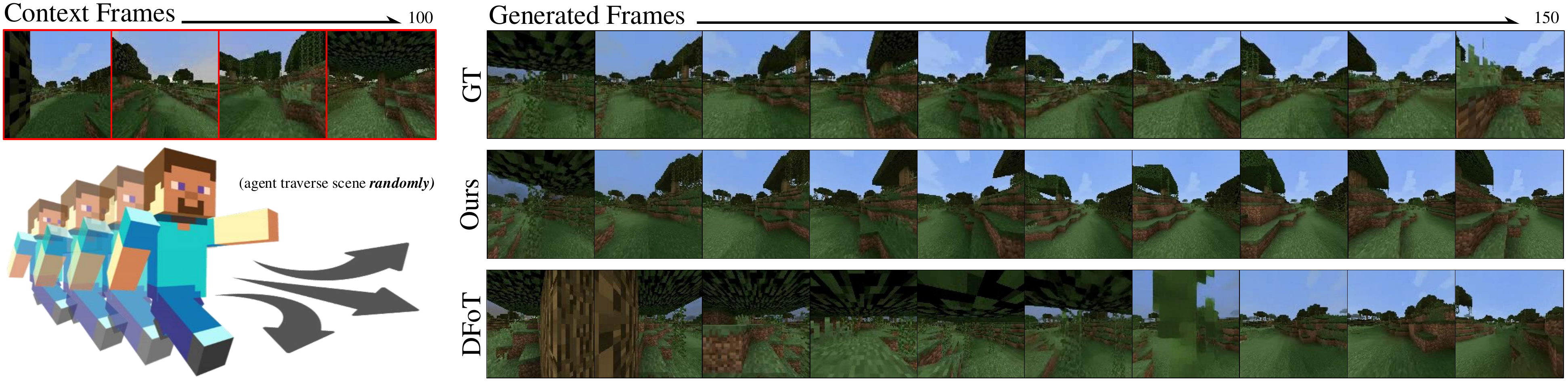}
    \caption{\textbf{Long memory video generation.} Our method generates sharp and consistent video predictions. Conditioned on agent actions, our method can accurately reconstruct previously visited regions of an environment, while maintaining linear training complexity and constant inference costs. In contrast, although state-of-the-art diffusion forcing transformers (DFoT~\cite{song2025history}) can generate consistent looking videos over long horizons, their memory is bounded by the maximum context seen during training. Unlike our method which has linear scaling, attention-based transformers training scales quadratically, making it prohibitively expensive to train on longer videos.}
    \label{fig:minecraft}
    \vspace{-4pt}
\end{figure*}

\vspace{-1em}
\paragraph*{Video world models.} 
World models learn to predict state transitions resulting from actions, enabling application in agent planning and interactive simulations.
Video prediction methods have been applied to learn visual world models.
%
While early research~\cite{ha2018recurrent,menapace2021playable,chiappa2017recurrent,oh2015action} typically relies on recurrent neural networks~(RNNs)~\cite{hochreiter1997long} and variational autoencoders~\cite{kingma2013auto}, recent works have shifted towards more scalable transformer-based diffusion/AR models.
%
%
Latest efforts in this direction have explored training such models without ground truth action data~\cite{Bruce2024GenieGI,menapace2021playable}, scaling to more complex worlds and controls~\cite{valevski2024diffusion,yang2023learning,xiang2024pandora,hu2023gaia}, enabling real-time inference~\cite{feng2024matrix,oasis2024}, and generating open-world environments~\cite{yu2025gamefactory,che2024gamegen,feng2024matrix}.
To support continuous generation, all these approaches evaluate their Transformer models in an autoregressive, sliding-window fashion, which inherently limits the model's context to only a few recent frames, typically covering at most a few seconds.
A direct consequence of this design is that these models inherently lack long-term coherence, a limitation explicitly pointed out by these works and directly experienced by practitioners~\cite{oasisdemo}.



\vspace{-1em}
\paragraph*{Linear attention.}

Recently, linear RNNs have been proposed as an efficient alternative to self-attention~\cite{vaswani2017attention}. The seminal work by \citet{katharopoulos2020transformers} introduced linear attention that replaces softmax in attention with a kernel function, which reduces runtime complexity from quadratic to linear.
%
Subsequent works have improved linear RNNs by refining the state update rule~\cite{schlag2021linear,behrouz2024titans,yang2024gated,daomamba2,gumamba}, enabling efficient parallelized~\cite{yang2025parallelizing} and hardware-aware training~\cite{daomamba2,yang2023gated}, or incorporating more expressive hidden states~\cite{sun2024learning,behrouz2024titans}.
One notable family of linear RNNs is state-space models (SSMs)~\cite{oshima2024ssm,gumamba,daomamba2}. SSMs can be interpreted as a hybrid of convolutional neural networks and recurrent neural networks that combines the best of both worlds—parallelizable training and efficient autoregressive inference.

Recent efforts have introduced SSMs into image and video generation domains by replacing self-attention layers with efficient Mamba layers~\cite{teng2024dim,liu2024linfusion,xie2024sana,yan2024diffusion,liu2024clear,wang2024lingen,gao2024matten}. These approaches typically perform bidirectional scanning over the entire token sequence, therefore not utilizing the efficient autoregressive inference capabilities of SSMs. 
In contrast, our work employs unidirectional SSMs for modeling temporal dynamics and world state transitions, naturally leveraging their inherent advantagess.
%


\section{Preliminaries}
\label{sec:background}

\subsection{Video Diffusion Models}
A diffusion process progressively corrupts an observed datapoint sampled from the data distribution $p(x_0)$ by adding Gaussian noise to it. The noisy data is given by:
\begin{equation}
x_t = \alpha_t x_0 + \sigma_t \epsilon,
\end{equation}
where the scalar parameters $\alpha_t, \sigma_t > 0$ control the signal-to-noise ratio according to a predefined noise schedule and $\epsilon$ is sampled from a standard Gaussian distribution. Diffusion models~\cite{sohl2015deep,ho2020denoising,song2021score} are typically trained to predict the noise by minimizing the following denoising objective:
\begin{equation}
    \mathcal{L}(\theta) = \mathbb{E}_{t, x_0, \epsilon}\left\|\epsilon_{\theta}(x_t, t) - \epsilon\right\|_2^2
    \label{eq:denoising_loss}
\end{equation}
or alternative but equivalent targets such as the original clean data $x_0$ or the velocity $\epsilon-x_0$. 

Diffusion models for video generation typically employ a two-stage approach: first encoding raw videos into latent space using a 3D variational autoencoder (VAE)~\cite{villegas2023phenaki,gupta2024photorealistic,videoworldsimulators2024}, then learning a diffusion model in this latent space. 

In conventional video diffusion models, the noise level is the same across all latent frames at each training iteration. This requires simultaneously generating all video frames during inference, where all frames following the same noise schedule. Diffusion forcing~\cite{chen2024diffusion} introduces a strategy where noise levels are sampled independently per frame during training, enabling sequential video generation at inference time by denoising each frame conditioned on previously generated clean frames. The ability to generate a video autoregressively conditioned on streaming controls is essential for world modeling applications, such as gaming or robotic learning. Denote $\{x_0^i\}_{i=1}^{T}$ as a sequence of $T$ latent frames. The noisy latent frames during training are obtained by
\begin{equation}
    x_{t_i}^i = \alpha_{t_i} x_0^i + \sigma_{t_i} \epsilon^i,
    \label{eq:diffusion_forcing}
\end{equation}
where $t_i$ is sampled independently for each frame $i$.
Previous approaches have explored various backbone architectures under the diffusion forcing training scheme, including recurrent neural networks~\cite{chen2024diffusion} and causal~\cite{yin2024causvid} or bidirectional~\cite{song2025history} transformers. 

The main component of the transformer architecture~\cite{vaswani2017attention} is self-attention, where each token in a sequence attends to all others via dot-product similarity between learned query, key, and value representations. Given a sequence of input embeddings \(X\), self-attention computes:
\begin{equation}
\text{Attn}(X) = \text{softmax} \left( \frac{QK^T}{\sqrt{d}} \right) V,
\end{equation}
where \(Q, K, V\) are linear projections of \(X\), and \(d\) is the latent dimension.
Causal attention is a variant of self-attention that only allows tokens to attend to previous tokens in the sequence. This is achieved by masking the attention matrix to prevent information flow from future tokens. In video diffusion models, previous approaches have applied block-wise causal mask to ensure that each token can only attend to tokens in the same or previous frames~\cite{yin2024causvid}, thereby enabling autoregressive generation with KV-caching.






\subsection{State-Space Models (SSMs)}

An SSM models a sequence dynamic as a linear system
\begin{equation}
    H_t = A H_{t-1} + B X_{t-1};\,\,\, X_t = C H_t + D X_{t - 1},
    \label{eq:plain_SSM}
\end{equation}
where $H_t$ are latent states, and $A$, $B$, $C$, $D$ are matrices with appropriate dimensions. 
Building on this framework, Mamba~\cite{gumamba} provides additional expressivity by modeling the parameters $A$, $B$, $C$, $D$ as linear projections of input $X_t$, \textit{i.e.,} $A_t := \text{Linear}_{\theta_A}(X_t)$ and similarly for $B_t$, $C_t$, and $D_t$.
This allows the sequence dynamic to be content-aware, and can be viewed as a generalization of causal linear attention, where
$B$ and $C$ play the role of $K$ and $Q$, respectively.

Unlike attention, however, the generation of new tokens during inference takes the form of expression (\ref{eq:plain_SSM}), and requires only the latent state $H_T$, where $T$ is the most recent token. 
This offers superior time and space complexity compared to attention (see Tab.~\ref{tab:model-comparison}), but at the cost of compressed memory representation, as inference no longer involves all-to-all comparison of past inputs.
%

\section{Methods}
\label{sec:method}

While SSMs have proven to be an effective drop-in replacements for attention in domains such as language modeling, directly substituting attention blocks with SSMs leads to suboptimal results for autoregressive video generation. In this section, we identify and address several shortcomings in the naive approach regarding \textbf{\textit{architecture}} (Sec.~\ref{subsec:arch}), and \textbf{\textit{training}} (Sec.~\ref{subsec:train}). Our solutions result in a model that displays long-term spatial memory while maintaining {\textbf{\textit{constant per-frame inference cost}} throughout generation (Sec.~\ref{subsec:infer}).}

\begin{figure}[t]   
    \centering
    \includegraphics[width=\linewidth]{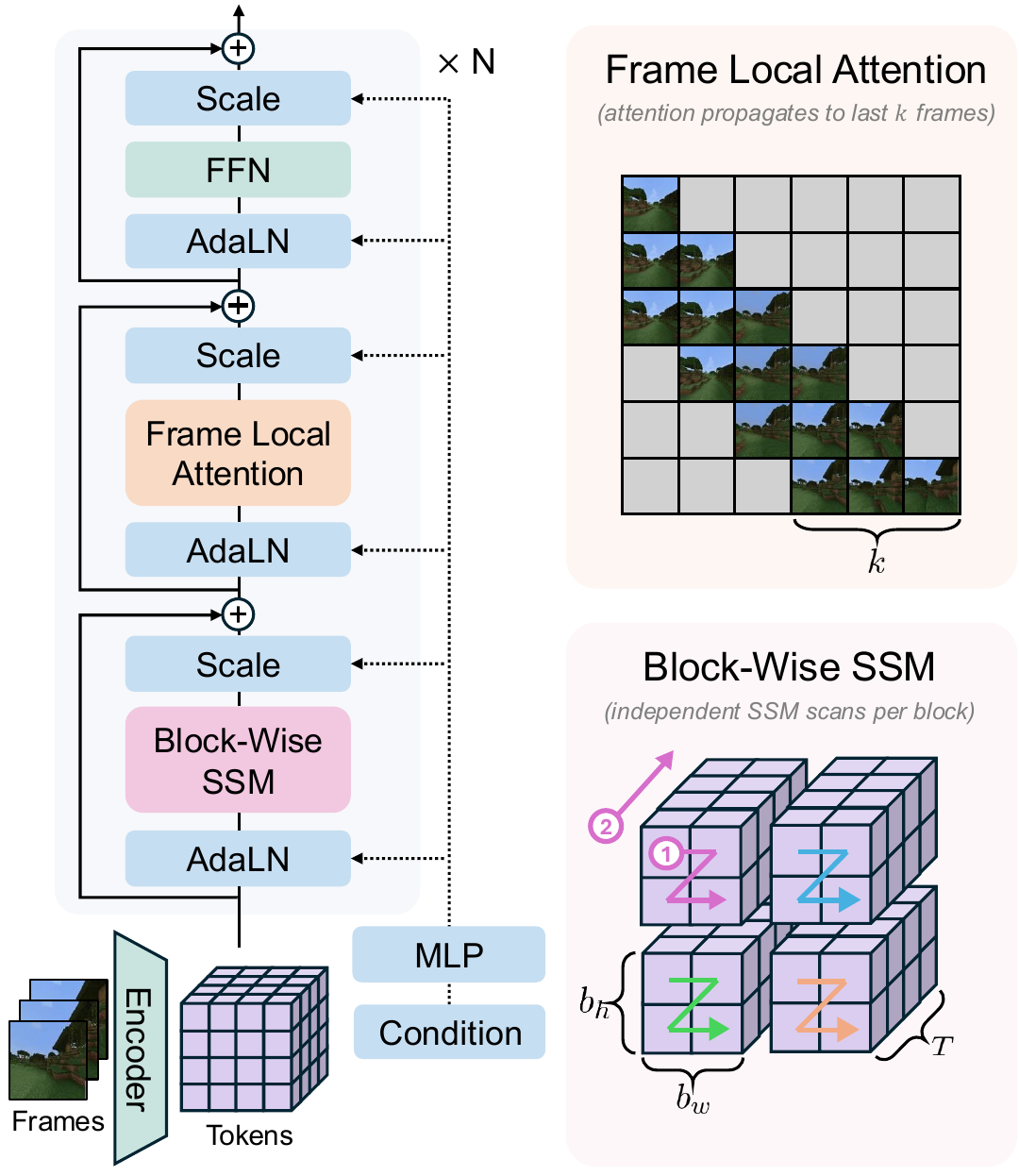}
    \caption{\textbf{Model architecture.} Our model features a block-wise SSM scan that divides spatial dimensions into independent scanning blocks $(b_h, b_w)$, balancing temporal memory with spatial coherence. This works alongside frame local attention, which enables bidirectional processing within frames while maintaining causal relationships across the previous $k$ frames, resulting in improved per-frame visual quality and temporal consistency. 
    }
    \label{fig:architecture}
    \vspace{-4pt}
\end{figure}

\subsection{Model Architecture}\label{subsec:arch}
Because our model generates video frames autoregressively (one frame at a time), the temporal dimension (the sequence of frames) must be placed at the end of the scanning order.
This ``spatial-major/time-minor'' ordering ensures that the model processes all spatial information within the current frame before moving to the next frame, thus preserving causal constraints and preventing the model from accessing future frame information.
However, the spatial-major scan order makes it challenging to capture long-term temporal dependencies, as temporally adjacent tokens become distant from each other in the flattened token sequence. To address this limitation, we introduce a method that balances temporal memory and spatial coherence using a block-wise reordering of the spatio-temporal tokens.

\vspace{-1em}
\paragraph*{Block-wise SSM scan.} As shown in Fig.~\ref{fig:architecture}~(bottom right), our method breaks up the original sequence of tokens along the spatial dimensions into blocks of size $(b_h, b_w, T)$, where $b_h$ and $b_w$ are layer-dependent block heights/width, and $T$ is the temporal dimension of the data. Instead of performing a single scan over the entire sequence of tokens, a separate scan is performed for each token block. By controlling the values of $b_h$ and $b_w$, we enable a trade-off between temporal correlation and spatial coherence. 
Temporally adjacent tokens are now separated by $b_h \times b_w$ tokens rather than $H \times W$~(as in conventional spatial-major scanning), where $H, W$ represent the height/width of each frame. However, smaller blocks lead to worse spatial coherence, as the independent scans prevent tokens in different blocks from interacting. Therefore, the choice of block-size represents an effective way of trading off consistent long-term memory for short-term spatial consistency. Our model leverages the benefits of both small and large block sizes, by employing different values for $b_h$ and $b_w$ in different layers.

\begin{figure}[t]   
    \centering
    \includegraphics[width=\linewidth]{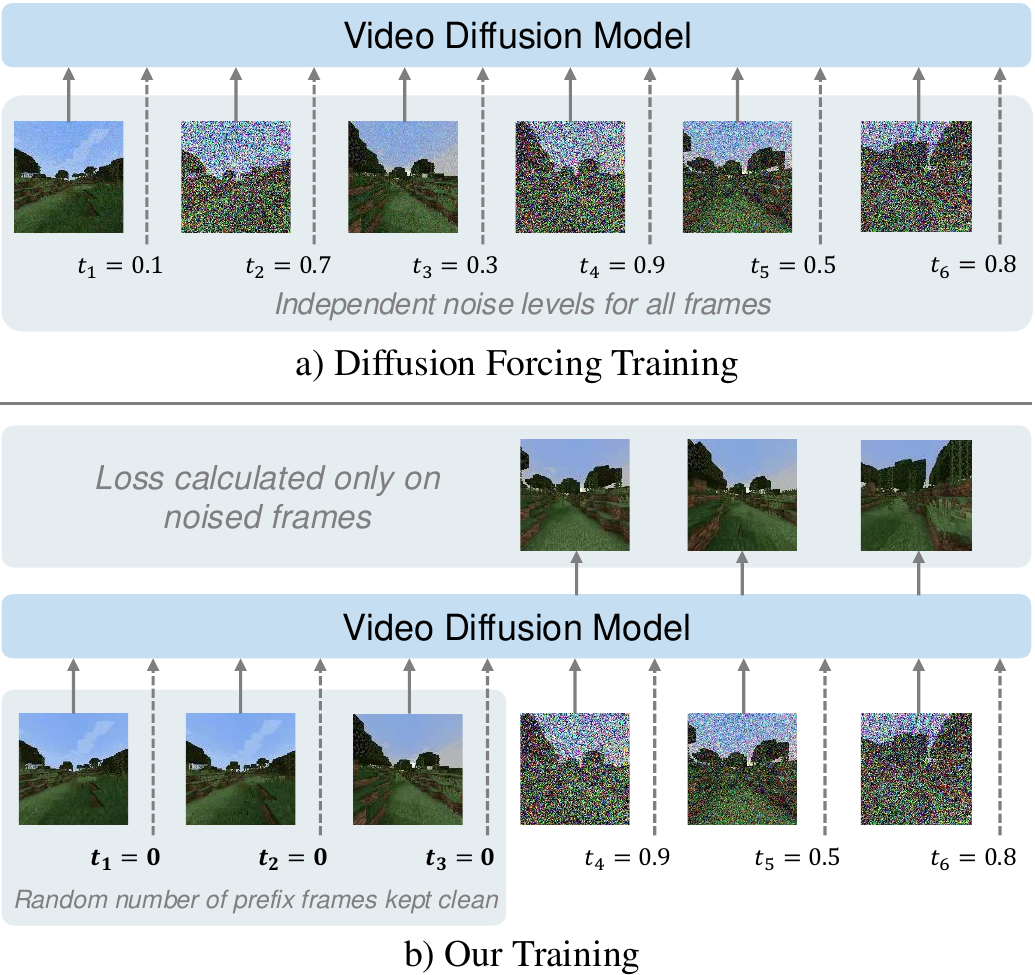}
    \caption{\textbf{Improved long-context training.} a) Standard diffusion forcing injects independent noise levels to all frames, b) Our method keeps a random number of initial frames completely clean ($t_i=0$), adds independent noise to later frames, and calculates loss only on the noised frames. By providing clean context frames distant from denoising targets, our approach encourages the model to learn long-term dependencies.}
    \label{fig:training}
    \vspace{-2pt}
\end{figure}

SSMs can struggle with high-complexity tasks such as visual generation due to the limited expressivity of the fixed-dimensional SSM state. Our block-wise scanning method mitigates this limitation by effectively increasing the dimensionality of the SSM state at each layer, as each block is allocated a separate state. 

\begin{figure*}[t!]   
    \centering
    \includegraphics[width=\linewidth]{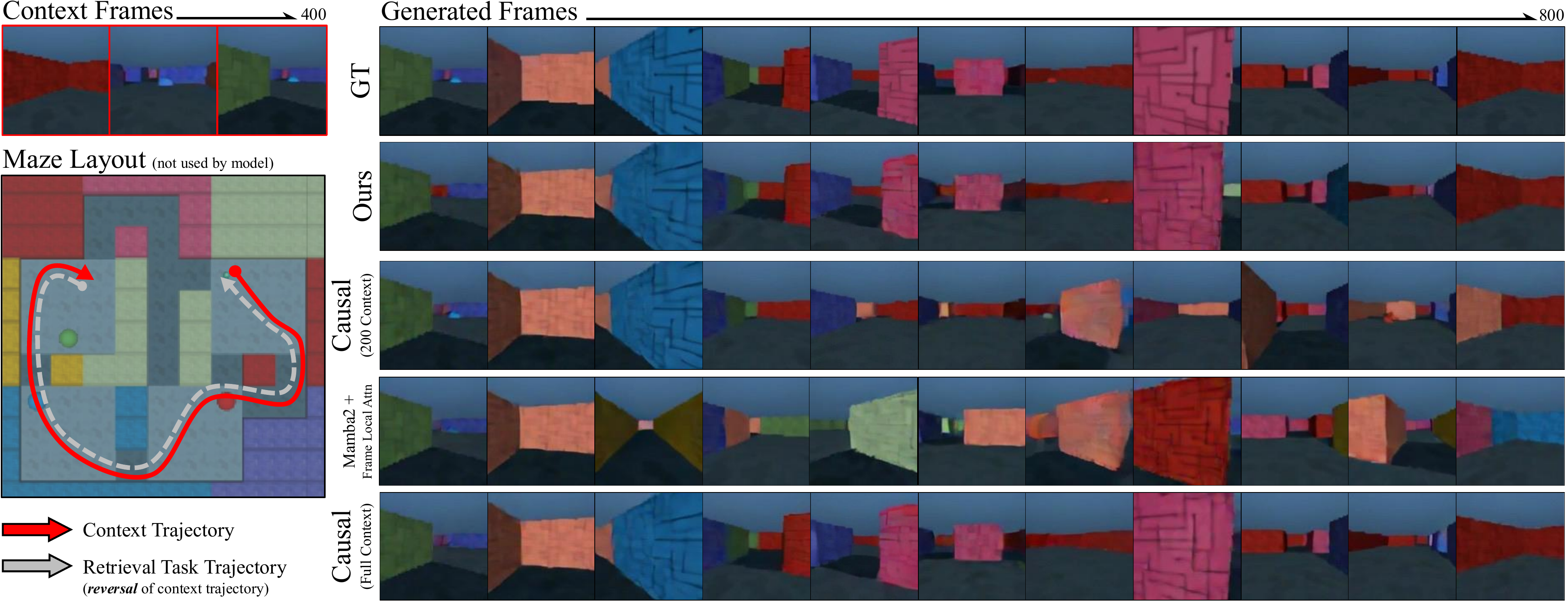}
    \caption{\textbf{Overview of retrieval task and qualitative results.} The maze layout shows the context trajectory (red) and retrieval trajectory (gray), which reverses the original path. We compare each model on a retrieval task with 400 generated frames following 400 context frames. Top row displays ground truth frames. Our model demonstrates high fidelity to ground truth throughout the sequence. The Causal model with 200-frame context deteriorates quickly beyond its training window, while Mamba2 + frame local attention fails completely. The Causal model with full context performs well but requires quadratic complexity during training and linear complexity during inference.}
    \label{fig:retrieval}
    \vspace{-2pt}
\end{figure*}

\vspace{-1em}
\paragraph*{Frame local attention.} Linear attention variants such as Mamba have shown to struggle in tasks related to associative recall~\cite{yang2024gated}. In video generation, the inability of Mamba to retrieve precise local information results in poor frame-wise quality and loss of short-term temporal consistency. Prior works~\cite{Ren2024SambaSH,yang2025parallelizing,yang2024gated,behrouz2024titans} have shown hybrid architectures that combine local attention with SSMs can improve language modeling. 
In our model, we introduce a frame-wise local attention block following every Mamba scan, as illustrated in Fig.~\ref{fig:architecture}. During training, we apply block-wise causal attention where each token can only attend to tokens in the same frame and a fixed-size window of previous frames. 
The attention mask $M$ takes the form,
\begin{equation}
M_{i,j} = 
\begin{cases}
1, & \text{if } j \in [i-k, i], \\
0, & \text{otherwise.}
\end{cases}
\label{eq:mask}
\end{equation}
where $i$ and $j$ are indices to frames in the sequence, and $k$ is the window size. 

\vspace{-1em}
\paragraph*{Action condition.} We enable interactive controls during autoregressive generation by passing actions corresponding to each frame as input. Continuous action values (e.g., camera positions) are processed through a small MLP and added to the noise level embeddings, which are then injected into the network via adaptive normalization layers~\cite{ba2016layer,huang2017arbitrary,huang2018multimodal,peebles2022scalable}. For discrete actions, we directly learn embeddings corresponding to each possible action.

\subsection{Long-Context Training}\label{subsec:train}
Although our architecture design enhances the model's ability to maintain long-term memory, it is still challenging to learn long temporal dependencies with standard diffusion training schemes.
Video data contains significant redundancy, allowing models to rely primarily on nearby frames for denoising in most cases. As a result, diffusion models frequently become trapped in local minima, failing to capture long-term dependencies.

Standard diffusion forcing always adds noise independently to each frame during training. Under these conditions, the model has limited incentive to reference distant context frames since they usually contain less useful information than local frames. To encourage the model to attend to distant frames and learn long-term correlations, we mix diffusion forcing with a modified training scheme that maintains a random-length prefix of frames completely clean (noise-free) during training, as illustrated in Fig.~\ref{fig:training}. When large noise is added to the later frames, the clean context frames may provide more useful information than the noisy local frames, prompting the model to utilize them effectively.
This is similar to the training scheme in Ca2-VDM~\cite{gao2024ca2} although our motivation is different and we still keep independent noise level for the later noisy frames.

\subsection{Efficient Inference via Fixed-Length State}\label{subsec:infer}

During inference, we autoregressively generate new video frames conditioned on input actions. Our hybrid architecture ensures constant speed and memory usage. Specifically, each layer of our model only tracks: (1) a fixed-length KV-cache for the previous $k$ frames, and (2) the SSM state for each block. This ensures constant memory usage throughout the generation process, unlike fully causal transformers whose memory requirements grow linearly as they store KV-caches for all previous frames. Similarly, our method maintains constant per-frame generation speed, as local attention and block-wise SSM computations do not scale with video length. This property is crucial for video world model applications, where generating video frames indefinitely without performance degradation is essential.


\section{Experiments}
\label{sec:experiments}

We evaluate our method in terms of training and inference efficiency, as well as long-term memory capabilities.
To this end, we utilize two long video datasets and evaluate the model's performance on spatial memory tasks that require recalling information from distant frames to generate accurate predictions. Details on the datasets are provided in Sec.~\ref{subsec:datasets}, metrics in Sec.~\ref{subsec:evaluations} and results in Sec.~\ref{subsec:results}.
%


\subsection{Datasets}
\label{subsec:datasets}
\paragraph*{Memory Maze.}
Memory Maze~\cite{pasukonis2022memmaze} is a 3D domain of randomized mazes,
designed for evaluating long-term memory capabilities of RL agents. Samples from the dataset contains trajectories collected from an agent exploring a randomly generated maze. Each trajectory contains 2000 action-frame pairs, which specify the most recent action/position of the agent, along with the observation of the maze from the agent's point-of-view.
We train all models on position/observation pairs. 
%
%
Any information regarding the ground truth layout of the maze is excluded.

\begin{figure*}[t]   
    \centering
    \includegraphics[width=\linewidth]{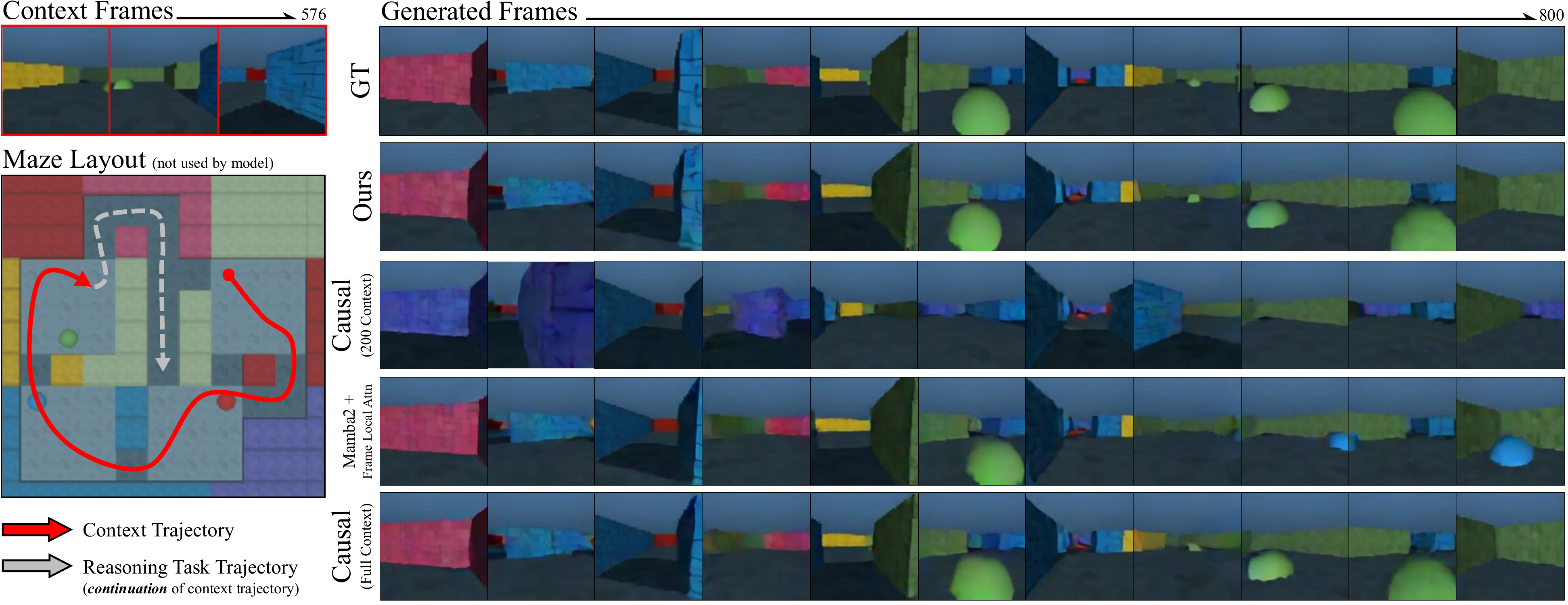}
    \caption{\textbf{Overview of reasoning task and qualitative results.} The maze layout illustrates the context trajectory (red) and reasoning trajectory (gray)\protect\footnotemark[1], which continues the context path. We compare each model on a retrieval task with 224 generated frames conditioned on 576 context frames.  The causal transformer trained on 192 frames fails immediately, as the queried region lies beyond its trained context. Mamba2 + Frame Local Attention fails to recall visual details such as positions of the balls. Our method successfully reconstructs previously visited regions of the maze, with performance comparable to the full-context causal transformer.}
    \label{fig:reasoning}
\end{figure*}

\vspace{-1em}
\paragraph{TECO Minecraft.} The TECO~\cite{Yan2022TemporallyCT} dataset consists of 200K gameplay trajectories collected from Minecraft. During data collection, an agent takes one of four actions (forward, turn left, turn right, jump) in random sequences, producing a trajectory consisting 150\footnotemark[1] action/observation pairs. These random action sequences result in trajectories where the agent occasionally revisit regions seen earlier. 
\footnotetext[1]{The original dataset contains 300 action/observation pairs, following~\cite{song2025history}, we sample every other frame.} 

\begin{table}[t!]
    \centering
    \small
    \begin{tabular}{lccc}
    \toprule
    \multirow{2}{*}{Model} & \multicolumn{3}{c}{Retrieval (400 Frames)} \\
    \cline{2-4}
     & SSIM $\uparrow$ & LPIPS $\downarrow$ & PSNR $\uparrow$ \\
    \hline
    Causal (192 Frame Context) & 0.829 & 0.147 & 26.4 \\
    Mamba2~\cite{daomamba2} & 0.747 & 0.313 & 20.4 \\
    Mamba2 + Frame Local Attn & 0.735 & 0.336 & 19.3 \\
    Ours &\textbf{0.898} & \textbf{0.069} & \textbf{30.8} \\
    \hdashline
    Causal (Full Context) & 0.914 & 0.057 & 32.6 \\
    \bottomrule
    \end{tabular}
    \caption{\textbf{Quantitative results on retrieval task.} Comparison of model performance on the 400-frame retrieval task using SSIM, LPIPS, and PSNR. Our model outperforms all baselines with sub-quadratic complexity, while approaching the performance of full-context causal transformers with quadratic training complexity.}
    \label{tab:retrieval}
    \vspace{-6pt}
\end{table}

\subsection{Evaluations}
\label{subsec:evaluations}
\paragraph*{Spatial retrieval task.} This task involves providing the model with a random agent trajectory and the corresponding observations as context, then tasking the model to backtrack through the exact sequence of actions to the agent's starting position. Given the scene is static, the generated sequence should reverse the context frames. We refer to this as a retrieval task because the ground truth answer can be retrieved from the given context frames. Fig.~\ref{fig:retrieval} shows an example of the context and generation trajectories. We evaluate the retrieval task only on the Maze dataset, as certain actions in Minecraft are not invertible (e.g. jump).

\vspace{-1em}
\paragraph*{Spatial reasoning task.} Similar to the retrieval task, the spatial reasoning task involves providing the model with a random agent trajectory and observations as context. However, instead of backtracking, the model continues the trajectory with random actions.
Assuming the model has been given enough context such that the entire environment has been committed into memory, the model should reconstruct every observation along the continued trajectory. Fig.~\ref{fig:reasoning} shows an example of the context and generated trajectories. 

\begin{table}[t!]
    \centering
    \small
    \begin{tabular}{lccc}
    \toprule
    \multirow{2}{*}{Model} & \multicolumn{3}{c}{Reasoning (224 Frames)} \\
    \cline{2-4}
     & SSIM $\uparrow$ & LPIPS $\downarrow$ & PSNR $\uparrow$ \\
    \hline
    Causal (192 Frame Context) & 0.839 & 0.125 & 27.1 \\
    Mamba2~\cite{daomamba2} &0.827 & 0.150 & 26.4  \\
    Mamba2 + Frame Local Attn &  0.845 & 0.113 & 27.5 \\
    Ours &\textbf{0.855} & \textbf{0.099} & \textbf{28.2} \\
    \hdashline
    Causal (Full Context) & 0.860 & 0.089 & 28.8 \\
    \bottomrule
    \end{tabular}
    \caption{\textbf{Quantitative results on reasoning task.} Performance comparison on the 224-frame reasoning task conditioned on 576 context frames. Our model surpasses all other sub-quadratic methods and performs close to full-context causal transformers.}
    \label{tab:reasoning}
    \vspace{-0.5em}
\end{table}



\subsection{Results} 
\label{subsec:results}
\paragraph*{Results on Memory Maze.} Tables~\ref{tab:retrieval} and~\ref{tab:reasoning} present quantitative comparisons in terms of spatial retrieval and reasoning on Memory Maze. 
We evaluate each model by comparing generated frames against ground truth using similarity metrics including SSIM~\cite{wang2004image}, LPIPS~\cite{zhang2018unreasonable}, and PSNR~\cite{huynh2008scope}.

For both tasks, we compare our method against baselines with sub-quadratic training complexity. These include a causal transformer trained on a limited context length\footnotemark[2], an architecture with only Mamba2 blocks, and Mamba2 + Frame Local Attn, a hybrid model that combines our frame local attention with Mamba2. We also include results from a causal transformer trained on the full context as reference.
Our model outperforms all sub-quadratic baselines across all metrics in both tasks. While the causal transformer with full context gives the best performance, it comes with quadratic training and inference complexity. As shown in Figures~\ref{fig:retrieval} and \ref{fig:reasoning}, frame predictions from other sub-quadratic models deviate from the ground truth after a certain period for both tasks, whereas our method maintains accurate predictions throughout the trajectory.
\footnotetext[1]{Trajectories shown in the figure are for illustrative purposes only. Actual trajectories tend to be longer, covers the whole maze, and revisits regions multiple times.}
\footnotetext[2]{Since the causal transformer has never seen the full context during training, we consider it sub-quadratic.}

\begin{figure}[t]   
    \vspace{-4pt}
    \centering
    \includegraphics[width=\linewidth]{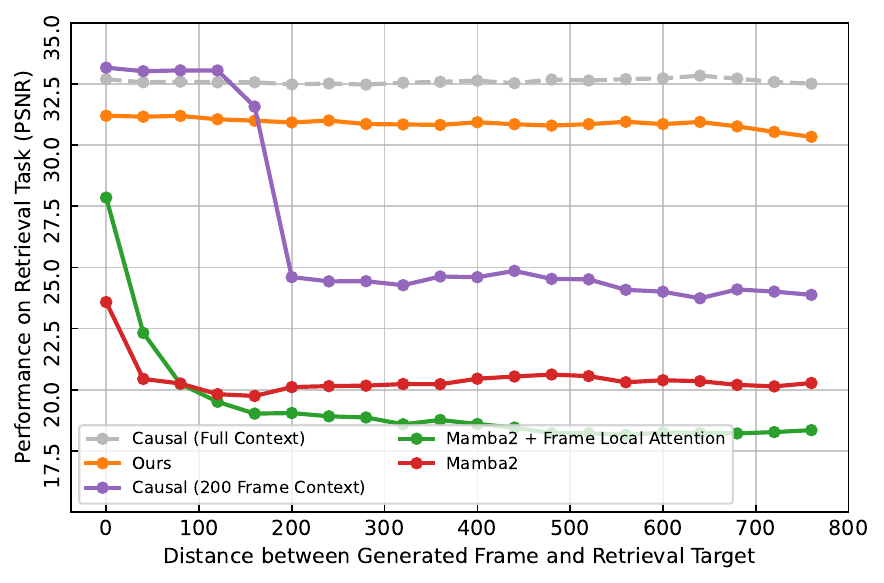}
    \caption{\textbf{Retrieval PSNR vs. Frame Distance.} Our model maintains consistent high performance comparable to full-context transformers while significantly outperforming limited-context transformers that degrade beyond training length and linear-complexity models that lack sufficient expressivity throughout.}
    \label{fig:retrieval_graph}
\end{figure}
\begin{figure*}[ht!]   
    \centering
    \includegraphics[width=\linewidth]{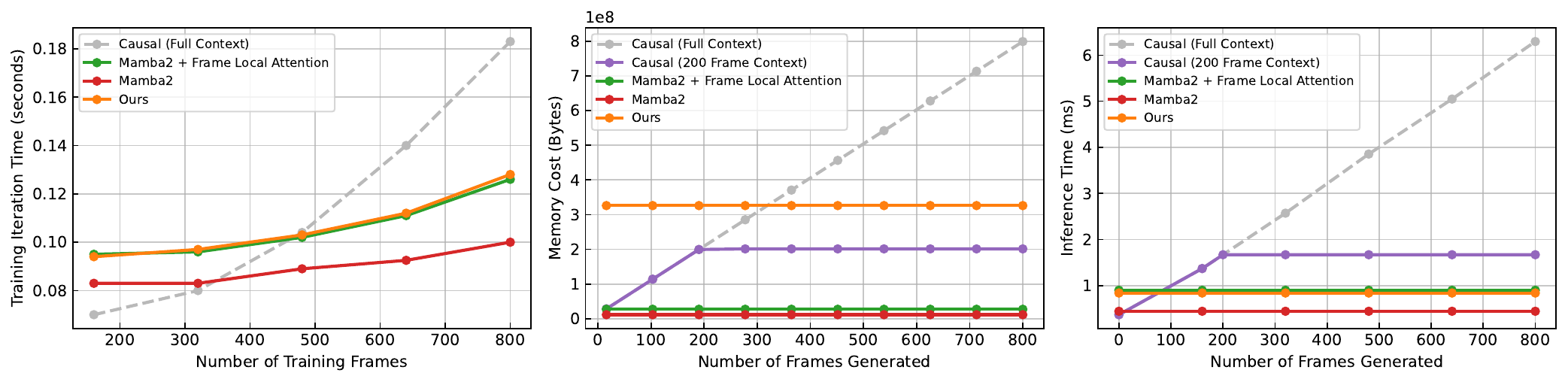}
    \caption{\textbf{Training and inference performance comparisons.} Evaluation of training costs (left), inference memory usage (center), and inference time (right), demonstrating how our approach maintains consistent memory and computational efficiency as frame count increases compared to baseline methods. 
    }
    \label{fig:costs}
    \vspace{-4pt}
\end{figure*}

In Fig.~\ref{fig:retrieval_graph}, we further analyze the performance of each method on the retrieval task, showing the change of retrieval accuracy as the distance between generated and retrieved frames increases. Causal transformers perform well within their training context but drop off quickly beyond their maximum training length. Other linear-complexity methods such as Mamba and Mamba2 + Frame Local Attn perform poorly due to limited state-space expressivity. In contrast, our method maintains high accuracy across all retrieval distances, comparable to a causal transformer trained on the full context.

\vspace{-1em}
\paragraph*{Results on TECO Minecraft.} We show quantitative and qualitative results on the Minecraft dataset in Tab.~\ref{tab:minecraft} and Fig.~\ref{fig:minecraft} respectively. We compare our method to diffusion forcing transformer~\cite{song2025history} (DFoT), a bidirectional transformer trained under the diffusion forcing regime. DFoT represents the current state-of-the-art architecture in autoregressive long video generation. However, due to the quadratic complexity of their model, DFoT is trained on a limited context length of 25 frames. As shown in Fig.~\ref{fig:minecraft}, our method can accurately predict previously explored regions, as opposed to methods with a limited context window.

Our method outperforms both DFoT and a causal transformer trained on a 25-frame context. All models obtain lower similarity on this dataset due to shorter trajectories, where the model is given only 100 frames of context to predict 50 frames.
Often, a 100-frame context is insufficient for the agent to fully observe the environment, causing task trajectories to venture into previously unseen regions in which case frame-wise similarity becomes less informative.

\begin{table}[t!]
    \vspace{2pt}
    \centering
    \small
    \begin{tabular}{lccc}
    \toprule
    \multirow{2}{*}{Model} & \multicolumn{3}{c}{Reasoning (50 Frames)} \\
    \cline{2-4}
     & SSIM $\uparrow$ & LPIPS $\downarrow$ & PSNR $\uparrow$ \\
    \hline
    DFoT~\cite{song2025history} & 0.450 & 0.281 & 17.1 \\
    Causal (25 Frame Context) & 0.417 & 0.350 & 15.8 \\
    Ours &\textbf{0.454} & \textbf{0.259} & \textbf{17.8} \\
    \bottomrule
    \end{tabular}
    \caption{\textbf{Reasoning task results on Minecraft dataset.} Performance comparison on the 50-frame reasoning task conditioned on 100 frames. Across all metrics, our model outperforms both DFoT~\cite{song2025history} and causal transformers limited to 25-frame context. 
    }
    \label{tab:minecraft}
\end{table}

\begin{table}[t!]
    \vspace{4pt}
    \centering
    \small
    \begin{tabular}{lccc}
    \toprule
    \multirow{2}{*}{Model} & \multicolumn{3}{c}{Reasoning (200 Frames)} \\
    \cline{2-4}
     & SSIM $\uparrow$ & LPIPS $\downarrow$ & PSNR $\uparrow$ \\
    \hline
    Ours w/o block-wise scan & 0.845 & 0.113 & 27.5 \\
    Ours with block size 1 & 0.766 & 0.198 & 23.1 \\
    Ours w/o Sec.~\ref{subsec:train} & 0.809 & 0.143 & 25.3 \\
    Ours (Full) &\textbf{0.855} & \textbf{0.099} & \textbf{28.2}  \\
    \bottomrule
    \end{tabular}
    \caption{\textbf{Ablations.} Performance of various setups of our model on the maze reasoning task. Every component, from architecture to training, is crucial for achieving accurate long-context memory.}
    \label{tab:ablations}
    \vspace{-2pt}
\end{table}

\vspace{-1em}
\paragraph*{Training and inference costs.} 
Fig.~\ref{fig:costs} evaluates model performance using three metrics: training cost per iteration~(left), memory utilization during generation~(center), and computational time during inference~(right). Our method demonstrates superior scaling across all metrics, with training time scaling linearly with context length while maintaining constant memory and computational costs during inference. For inference runtime comparison, we compare the runtime of a single forward pass through our frame local attention plus SSM update against full attention with KV-caching across all previously generated frames. 

\subsection{Ablations.}
\paragraph*{Block-wise SSM scan.} Tab.~\ref{tab:ablations} shows the benefits of our block-wise SSM scan. Without it and with only a single scan along all spatiotemporal tokens, the model struggles with maintaining memory over long horizons due to the lack of proximity in adjacent temporal tokens and limited SSM state capacity. Conversely, always using the smallest possible block size (i.e., $b_h,b_w=1$) ensures temporal token proximity but sacrifices spatial correlations, resulting in poor performance on the reasoning task where temporal retrieval alone is insufficient.

\vspace{-1em}
\paragraph*{Long-context training.}Tab.~\ref{tab:ablations} highlights the importance the training scheme outlined in Sec.~\ref{subsec:train}. Without this adjustment to diffusion forcing, our model falls into a local minimum and generates frames without looking at distant context, resulting in poor spatial reasoning performance.
\section{Limitations and Future Work}
\label{sec:discussion}
While our work represents a significant step towards scalable and consistent world models, it has certain limitations. 
First, despite achieving constant inference time, our method does not yet support interactive frame rates. Future work could speed-up the generation through timestep distillation~\cite{yin2024causvid}.
Second, our method cannot effectively handle memory longer than the training context length. Recent works~\cite{ben2024decimamba,yelongmamba} on length extrapolation for Mamba architectures could potentially extend our memory capacity.
Finally, our experiments are limited to low-resolution synthetic videos due to computational constraints. Scaling up to high-resolution, realistic videos is left for future work.


\newpage
{
    \small
    \bibliographystyle{ieeenat_fullname}
    \bibliography{main}
}

\clearpage
\setcounter{section}{0}
\setcounter{figure}{0}
\setcounter{table}{0}
\renewcommand{\thesection}{S\arabic{section}}
\renewcommand{\thefigure}{S\arabic{figure}}
\renewcommand{\thetable}{S\arabic{table}}




%
\definecolor{iccvblue}{rgb}{0.21,0.49,0.74}
\definecolor{moderate}{rgb}{0.85, 0.55, 0.1} 
\definecolor{BrightRed}{rgb}{0.85, 0.25, 0.25}
\definecolor{ViridianGreen}{rgb}{0.25, 0.6, 0.45}


\def\paperID{7620} 
\def\confName{ICCV}
\def\confYear{2025}



\maketitle

\section{Additional Results.}
\begin{figure*}[ht!]   
    \centering
    \includegraphics[width=\linewidth]{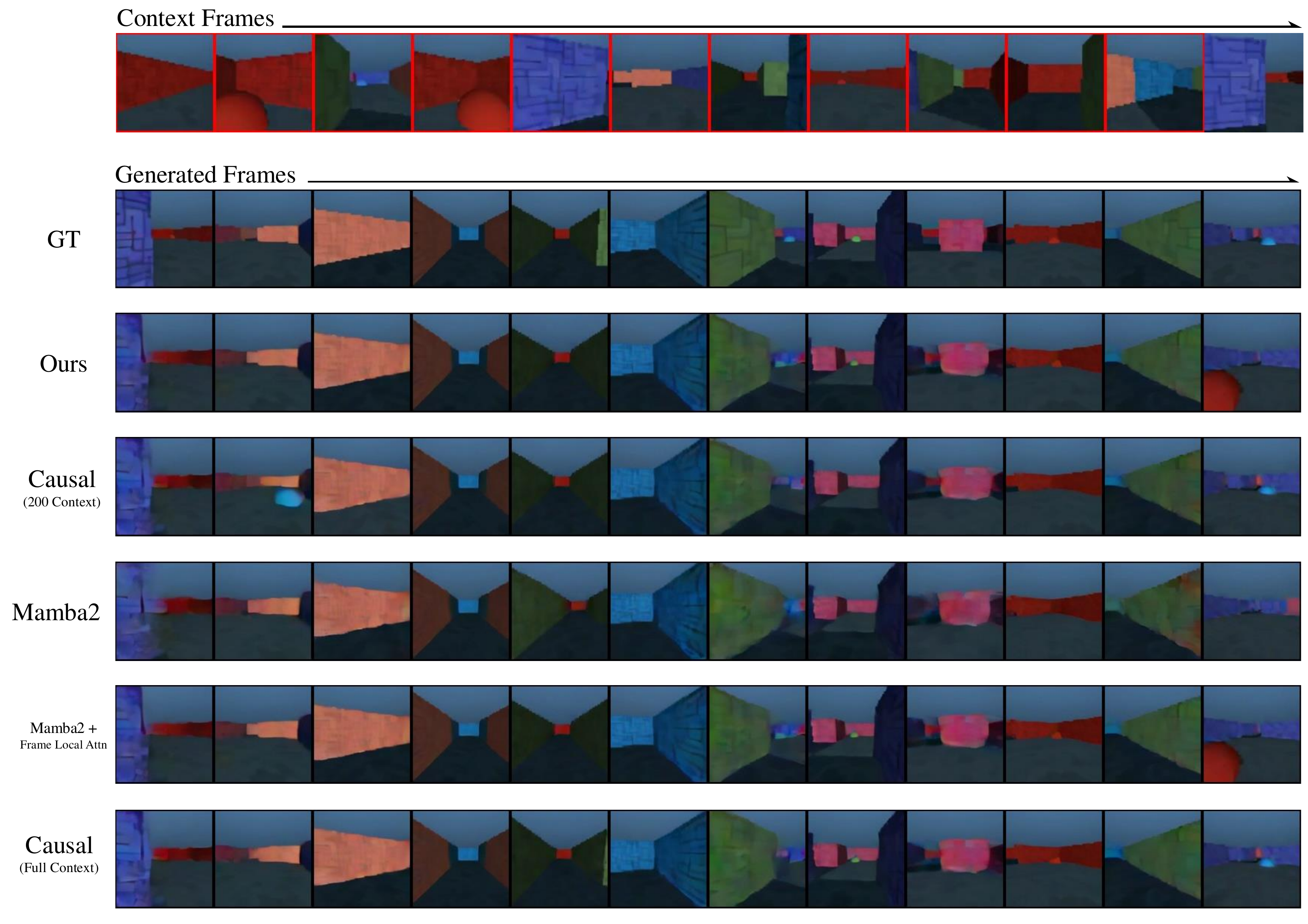}
    \caption{Additional results on reasoning task for the maze dataset.
    }
    \label{fig:supp_reasoning}
    \vspace{-4pt}
\end{figure*}

\begin{figure*}[ht!]   
    \centering
    \includegraphics[width=\linewidth]{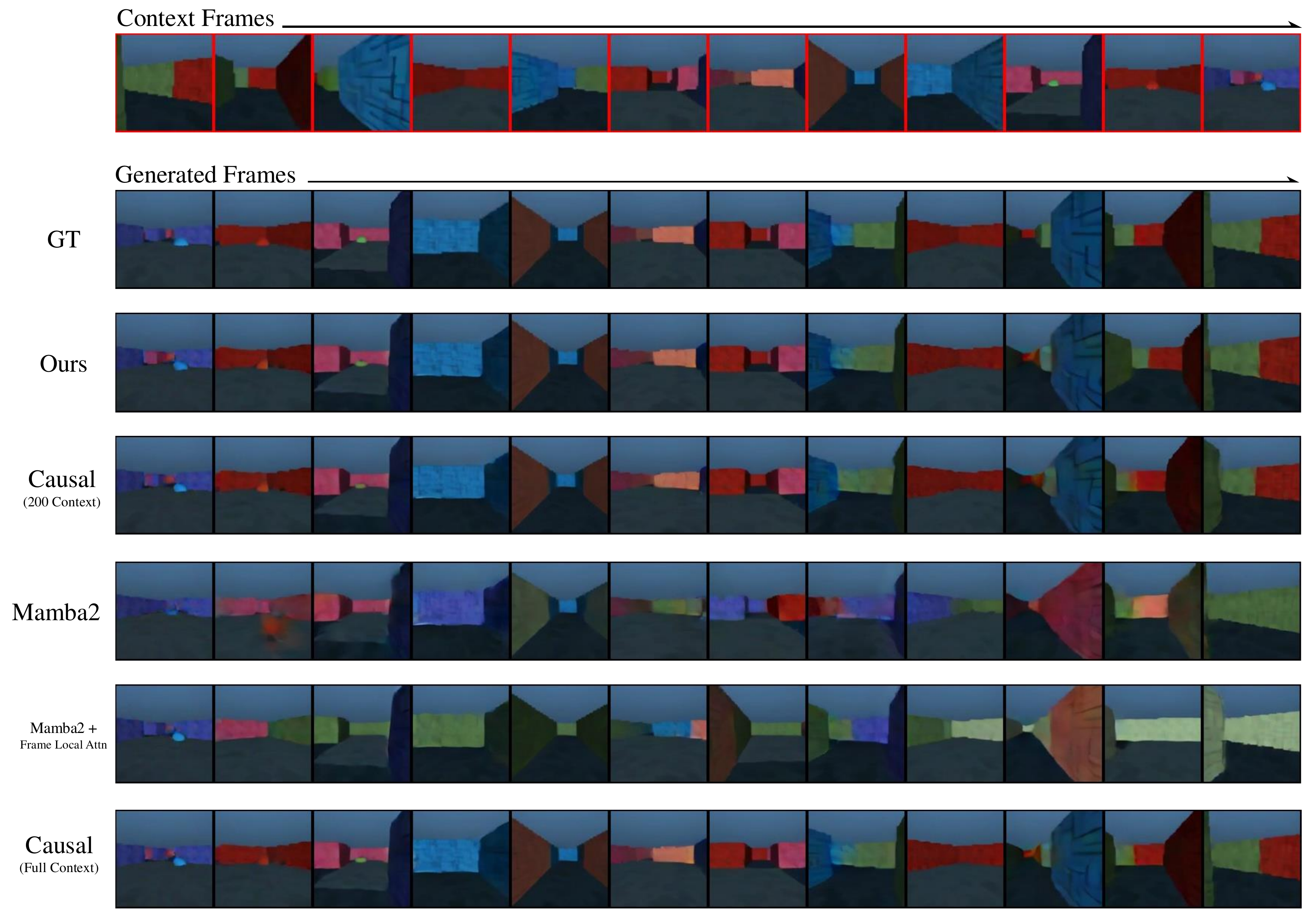}
    \caption{Additional results on retrieval task for the maze dataset.
    }
    \label{fig:supp_retrieval}
    \vspace{-4pt}
\end{figure*}

\begin{figure*}[ht!]   
    \centering
    \includegraphics[width=\linewidth]{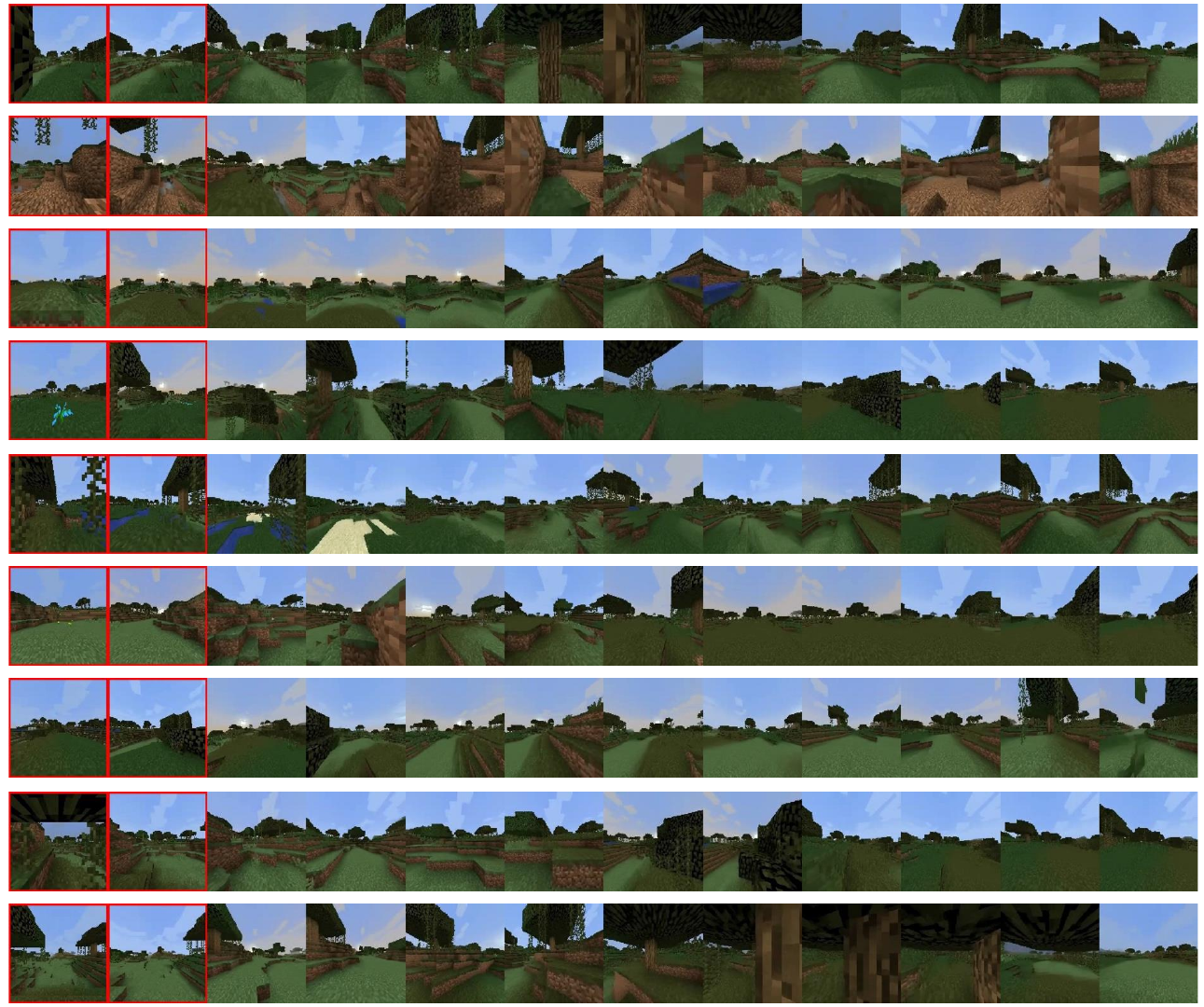}
    \caption{Additional results on long videos generated with our model trained on the minecraft dataset. Model if given 25 context frames and 125 random actions. Context frames are highlighted by a red border. Frames are sampled evenly from from all 150 output frames (including context).
    }
    \label{fig:supp_mc}
    \vspace{-4pt}
\end{figure*}

We show additional results for both spatial reasoning and retrieval tasks for the maze dataset in Supp. Fig.~\ref{fig:supp_reasoning} and ~\ref{fig:supp_retrieval}. We additionally include results of our method generating long-horizon videos conditioned on short context in Supp. Fig.~\ref{fig:supp_mc}. These results are best viewed as videos, we encourage readers to refer to the website attached to the supplementary materials packet. 

Additionally, we include FVD results of comparing our method against all relevant baselines under a long-horizon generation task of 240 context frames and 560 generated frames. As shown in Tab.~\ref{tab:fvd}, our method achieves the lowest FVD, even compared to a causal transformer trained on the entire context.

\begin{table}[h]
    \centering
    \small
    \begin{tabular}{lc}
    \toprule
    Method & FVD (240 + 560) $\downarrow$ \\
    \hline
    Causal (192 Frame Context) & 78.9 \\
    Causal (Full Context) & 45.1 \\
    Mamba2 & 163 \\
    Mamba2 + Frame Local Attn & 45.8 \\
    Ours & \textbf{38.9} \\
    \bottomrule
    \end{tabular}
    \caption{\textbf{FVD scores for long-term generation.} Comparison of Fréchet Video Distance (FVD) scores over 560 generated frames given 240 context frames. Our method achieves the lowest FVD, outperforming even causal with full context.}
    \label{tab:fvd}
\end{table}

\section{Implementation Details.}

\paragraph*{Latent diffusion.} Due to the dense information carried in videos, we trained our models on encoded versions of the data. For the Maze dataset, due to the large amount of frames, we use an internal VAE, which compresses token both spatially and temporally. For the minecraft dataset, due to the low number of raw frames, and fair comparisons, we use the same image VAE as DFot~\cite{song2025history}.

\paragraph*{Models.} We employ an architecture similar to CogVideo-X~\cite{yang2024cogvideox}. Each baseline model is built upon the same model architecture, opting to replace the attention blocks in each model block with the relevant mechanisms. Depending on the model, the parameter count per layer differs. For fairness, we kept parameter counts for all baselines and comparisons at 200M by adjusting the number of layers for each model. 

\paragraph*{Training.} We train models for different number of iterations depending on task and dataset. For the maze reasoning task, we the first train our model on videos with 400 frames for 150K iterations, then fine-tune this model on 800 frame videos for another 250K steps. Similarly, for the maze retrieval task, we train the model for 100K steps on the 400 frame videos, and 50K extra steps on 800 frames videos. For the minecraft model, we trained on the full 300 frames for 100K steps. We employ our long-context training regime during all stages and for all models. Using a ratio of $p=0.5$, we sample a random length prefix of the frame sequence to keep un-noised. The length of the prefix must exceed half of the total length of the training sequence to further encourage long-context training. When we don't sample a prefic, we keep all tokens noised, in this case, training is the same as diffusion forcing. Note that diffusion forcing is a special case of long-context training, when prefix length is zero.

\paragraph*{Frame local attention.} We observed significant speedup when using our frame local attention, compared to a fully causal mask by utilizing FlexAttention~\cite{Dong2024FlexAA}. For all of our experiments, we chose a frame window size of $k=10$. For faster training and sampling speeds, we group frames into chunks of 5. In our implementation of frame local attention, frames in a chunk maintain bidirectionality, while also paying attention to frames in the previous chunk, making the effective frame window 10.

\newpage

\end{document}